\documentclass[letterpaper, 10 pt, conference]{ieeeconf}  %

\IEEEoverridecommandlockouts                              %

\let\labelindent\relax

\usepackage{multicol}

\makeatletter
\let\NAT@parse\undefined
\makeatother

	\usepackage{amsmath}
	\usepackage{amsfonts}
	\usepackage{amssymb}
	\usepackage{ragged2e}
	\usepackage{graphicx}
	\usepackage{cancel}
	\usepackage{mathtools}
	\usepackage{tabularx}
	\usepackage{arydshln}
	\usepackage{tensor}
	\usepackage{array}
	\usepackage[dvipsnames]{xcolor}
	\usepackage{listings}
	\usepackage{textcomp}
	\usepackage{mathrsfs}
	\usepackage{bbm}
	\usepackage{tikz}
	\usepackage{tikz-cd}
	\usepackage{enumitem}
	\usepackage{arydshln}
	\usepackage{relsize}
	\usepackage{multirow}
	\usepackage{scalerel}
	\usepackage{upgreek}
	\usepackage{ifthen}
	\usepackage{yhmath}
	\usepackage{blkarray}
	\usepackage{dashrule}
	\usepackage{subcaption}
	\usepackage{overpic}

		\newcommand{\inv}{^{-1}}
		\newcommand{\abs}[1]{\left|#1\right|}

		\newcommand{\of}{\circ}

		\newcommand{\set}[1]{\left\{#1\right\}}

		\newcommand{\norm}[1]{\abs{\abs{#1}}}
		\newcommand{\tpose}{^{T}}

		\DeclareMathOperator{\argmin}{argmin}

		\newcommand{\st}{s.t.}

		\newcommand{\ptxt}[1]{\textrm{\textnormal{#1}}}
		\newcommand{\mf}[1]{\mathfrak{#1}}
		\newcommand{\mc}[1]{\mathcal{#1}}
		
		\newcommand{\mb}[1]{\mathbb{#1}}
		
		\DeclareMathAlphabet{\mathsfit}{T1}{\sfdefault}{\mddefault}{\sldefault}

	\usepackage{letltxmacro}
	\LetLtxMacro\orgvdots\vdots
	\LetLtxMacro\orgddots\ddots

	\makeatletter
	\DeclareRobustCommand\vdots{%
		\mathpalette\@vdots{}%
	}
	\newcommand*{\@vdots}[2]{%
		\sbox0{$#1\cdotp\cdotp\cdotp\m@th$}%
		\sbox2{$#1.\m@th$}%
		\vbox{%
			\dimen@=\wd0 %
			\advance\dimen@ -3\ht2 %
			\kern.5\dimen@
			\dimen@=\wd2 %
			\advance\dimen@ -\ht2 %
			\dimen2=\wd0 %
			\advance\dimen2 -\dimen@
			\vbox to \dimen2{%
				\offinterlineskip
				\copy2 \vfill\copy2 \vfill\copy2 %
			}%
		}%
	}
	\DeclareRobustCommand\ddots{%
		\mathinner{%
			\mathpalette\@ddots{}%
			\mkern\thinmuskip
		}%
	}
	\newcommand*{\@ddots}[2]{%
		\sbox0{$#1\cdotp\cdotp\cdotp\m@th$}%
		\sbox2{$#1.\m@th$}%
		\vbox{%
			\dimen@=\wd0 %
			\advance\dimen@ -3\ht2 %
			\kern.5\dimen@
			\dimen@=\wd2 %
			\advance\dimen@ -\ht2 %
			\dimen2=\wd0 %
			\advance\dimen2 -\dimen@
			\vbox to \dimen2{%
				\offinterlineskip
				\hbox{$#1\mathpunct{.}\m@th$}%
				\vfill
				\hbox{$#1\mathpunct{\kern\wd2}\mathpunct{.}\m@th$}%
				\vfill
				\hbox{$#1\mathpunct{\kern\wd2}\mathpunct{\kern\wd2}\mathpunct{.}\m@th$}%
			}%
		}%
	}
	\makeatother

	\tikzset{
	  symbol/.style={
		draw=none,
		every to/.append style={
		  edge node={node [sloped, allow upside down, auto=false]{$#1$}}}
	  }
	}

\usepackage[bookmarks=true]{hyperref}

\usepackage{cleveref}
\Crefname{figure}{Figure}{Figures}
\Crefname{table}{Table}{Tables}
\Crefname{equation}{Eq.}{Eqs.}
\Crefname{section}{Section}{Sections}
\Crefname{subsection}{Subsection}{Subsections}
\Crefname{appendix}{Appendix}{Appendices}

\DeclareMathOperator{\SE}{SE}

\DeclareMathOperator{\VALID}{VALID}
\DeclareMathOperator{\FREE}{FREE}
\newcommand{\leftarm}{\mathbf{L}}
\newcommand{\rightarm}{\mathbf{R}}
\newcommand{\CS}{\textsf{CS}}

\usepackage[commentColor=black]{algpseudocodex}
\tikzset{algpxIndentLine/.style={draw=black}}
\algrenewcommand{\alglinenumber}[1]{\bfseries\footnotesize #1}
\algrenewcommand{\textproc}{}
\algrenewcommand{\algorithmicrequire}{\textbf{Input:}}
\algrenewcommand{\algorithmicensure}{\textbf{Output:}}
\usepackage{algorithm}
\makeatletter
\newcommand{\algorithmname}{\ALG@name}
\renewcommand{\floatc@ruled}[2]{{\@fs@cfont #1:} #2\par}
\makeatother

\usepackage{color-edits}

\addauthor{tc}{cyan}
\addauthor{ms}{magenta}
\addauthor{ss}{teal}

\title{\LARGE \bf
Constrained Bimanual Planning with Analytic Inverse Kinematics
}

\author{
Thomas Cohn, Seiji Shaw, Max Simchowitz, and Russ Tedrake
\thanks{This work was supported by Amazon.com, PO No. 2D-06310236, the MIT Quest for Intelligence, and the National Science Foundation Graduate Research Fellowship Program under Grant No. 2141064. Any opinions, findings, and conclusions or recommendations expressed in this material are those of the author(s) and do not necessarily reflect the views of the National Science Foundation. The authors are with the Computer Science and Artificial Intelligence Laboratory (CSAIL), Massachusetts Institute of Technology, Cambridge, Massachusetts {\tt\small [tcohn,seijis,msimchow,russt]@mit.edu}}%
}

\begin{document}

\maketitle
\thispagestyle{empty}
\pagestyle{empty}

\begin{abstract}
In order for a bimanual robot to manipulate an object that is held by both hands, it must construct motion plans such that the transformation between its end effectors remains fixed.
This amounts to complicated nonlinear equality constraints in the configuration space, which are difficult for trajectory optimizers.
In addition, the set of feasible configurations becomes a measure zero set, which presents a challenge to sampling-based motion planners.
We leverage an analytic solution to the inverse kinematics problem to parametrize the configuration space, resulting in a lower-dimensional representation where the set of valid configurations has positive measure.
We describe how to use this parametrization with existing motion planning algorithms, including sampling-based approaches, trajectory optimizers, and techniques that plan through convex inner-approximations of collision-free space.

\end{abstract}

\section{Introduction}
\label{sec:introduction}

Enabling bimanual robots to execute coordinated actions with both arms is essential for achieving (super)human-like skill in automation and home contexts.
There exists a variety of tasks that are only solvable when two arms manipulate in concert \cite{krebs2022bimanual}, such as carrying an unwieldy object, folding clothes, or assembling parts.
In many manipulation tasks, one gripper can be used to provide fixture to the manipuland, while the other performs the desired action \cite{holladay2024robust}; such tasks include opening a bottle, chopping vegetables, and tightening a bolt.
Furthermore, some tools explicitly require two arms to use, such as hand mixers, rolling pins, and can openers.

To accomplish many of these desired tasks, the motion of the robot arms becomes subject to equality constraints imposed in task space.
For example, when moving an object that is held by both hands, the robot must ensure that the transformation between the end effectors remains constant.
Such task space constraints appear as complicated nonlinear equality constraints in configuration space, posing a major challenge to traditional motion planning algorithms.

\begin{figure}
	\centering
	\includegraphics[width=\linewidth]{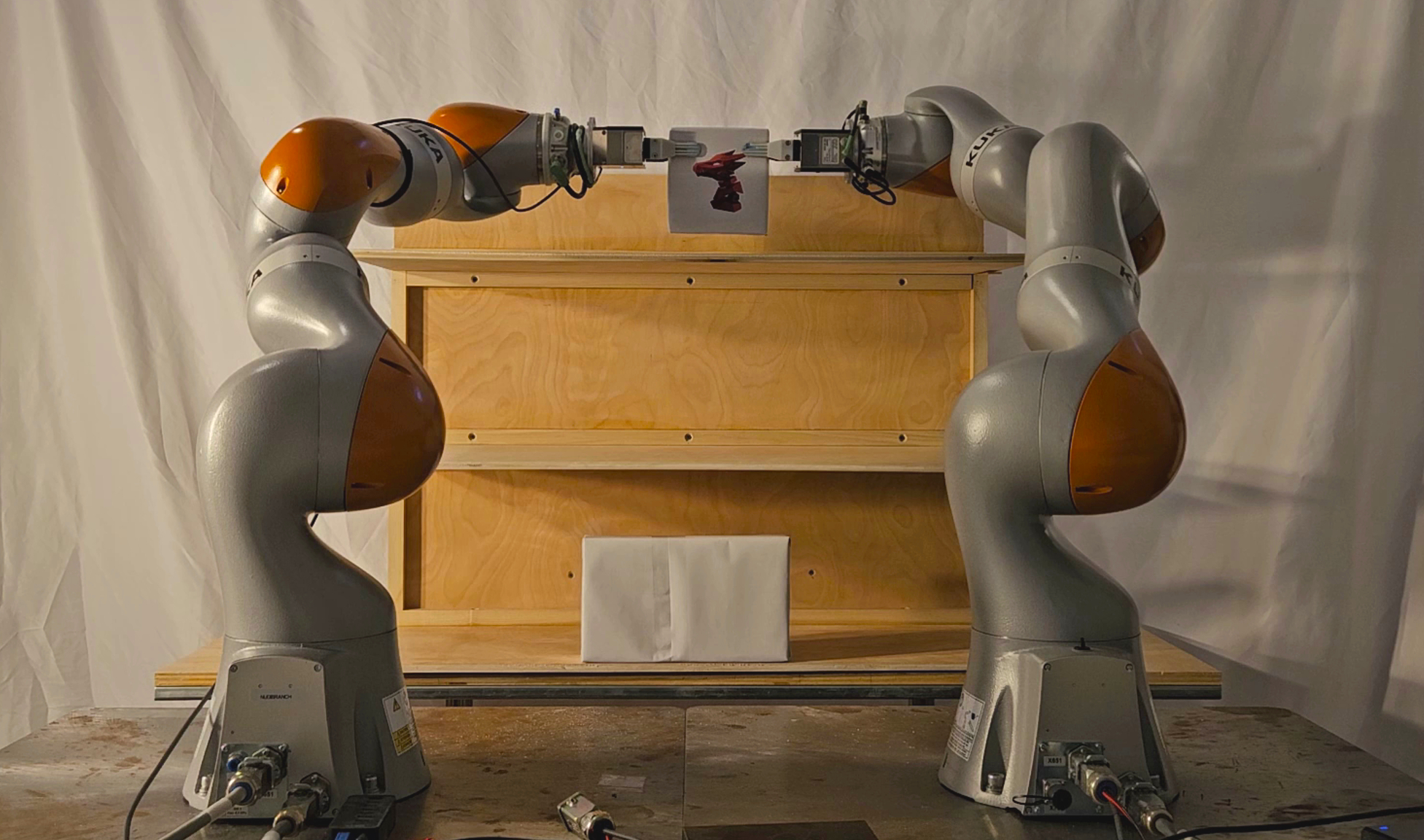}
	\caption{
		Hardware setup for our experiments.
		The two arms must work together to move an objects between the shelves, avoiding collisions and respecting the kinematic constraint.
	}
	\label{fig:teaser}
\end{figure}

In the existing literature, there are general techniques for handling task-space constraints in configuration-space planning.
Sampling-based planners can project samples onto the constraint manifold~\cite{berenson2009manipulation} or use numerical continuation~\cite{krauskopf2007numerical} to construct piecewise-linear approximations.
Constraints can also be relaxed~\cite{bonilla2015sample} or enforced directly with trajectory optimization~\cite{pavone2019trajectory}.
In the case of certain bimanual planning problems, there is additional structure that is not exploited by these general methods.
For certain classes of robot arms, \emph{analytic inverse kinematics} (analytic IK) can be used to map an end-effector pose (along with additional parameters to resolve kinematic redundnacy) to joint angles in closed form.
Such solutions are specific to certain classes of robot arms, but are a powerful tool to be leveraged if available.
Fortunately, analytic IK \emph{is} available for many popular robot arms available today, including the KUKA iiwa. See \Cref{fig:teaser}.

If a robot must move an object that it is holding with both hands, we propose constructing a plan for one ``controllable'' arm, and then the other ``subordinate'' arm can be made to follow it via an analytic IK mapping.
Configurations where the subordinate arm cannot reach the end-effector of the primary arm, or where doing so would require violating joint limits, are treated as obstacles.
In this way, we parametrize the constraint manifold so that the feasible set has positive measure in the new planning space.
Because we no longer have to consider the equality constraints, sampling-based planning algorithms can be applied without modification.
We can also differentiate through the IK mapping, enabling the direct application of trajectory optimization approaches.

The remainder of this paper is organized as follows.
First, we give an overview of the existing techniques used for constrained motion planning, and describe the available analytic IK solutions.
Then, we present our parametrization of the constraint manifold for bimanual planning, and discuss its relevant geometric and topological properties.
We describe the slight modifications which are necessary to adapt standard planning algorithms (including sampling-based planning and trajectory optimization) to operate in this framework.
We then present a technique for generating convex sets in this new configuration space, such that every configuration within such a set is collision free and kinematically valid.
These sets are essential for planning frameworks such as Graph of Convex Sets (GCS)~\cite{marcucci2024shortest,marcucci2023motion}.
Finally, we present various experiments demonstrating the efficacy of these new techniques.

\section{Related Work}
\label{sec:related_work}

Constrained bimanual planning is an instance of the general problem of motion planning with task-space constraints, a well-studied problem in robotics.
There has been extensive research into sampling-based approaches; these techniques fall under a number of broad categories:
\begin{itemize}
	\item Relax the constraints (with real or simulated compliance) to give the feasible set nonzero volume~\cite{bonilla2015sample,bonilla2017noninteracting}.
	\item Project samples to the constraint manifold~\cite{berenson2009manipulation,stilman2010global,mirabel2016hpp}.
	\item Construct piecewise-linear approximations of the constraint manifold~\cite{jaillet2012path,bohigas2013planning,kim2016tangent,kingston2019exploring}.
	\item Parametrize of the constraint manifold to eliminate constraints~\cite{han2008convexly,mcmahon2018sampling}.
	\item Build offline approximations of the constraint manifold, to simplify online planning~\cite{zha2018learning,csucan2012motion}.
\end{itemize}
See the survey paper~\cite{kingston2018sampling} for a more detailed summary.

Beyond sampling-based planning, standard nonconvex trajectory optimization approaches can handle arbitrary constraints, although they will generally only converge to a feasible solution with good initialization~\cite{kingston2018sampling}.
\cite{pavone2019trajectory} used sequential convex programming on manifolds for nonconvex trajectory optimization.
\cite{bordalba2022direct} greatly reduced constraint violations when computing trajectories on manifolds by enforcing collocation constraints in local coordinates.

Other approaches are designed specifically for the constraints that arise from robot kinematics.
Inverse kinematics (IK) -- computing robot joint angles so as to place the end effector at a given configuration -- has been especially applicable.
IK has long been used to sample constraint-satisfying configurations for bimanual robots, enabling the use of sampling-based planning algorithms~\cite{han2001kinematics,cortes2005sampling,gharbi2008sampling,wang2019inverse}.
IK can be leveraged to find stable, collision-free configurations for a humanoid robot~\cite{kanehiro2012efficient}, to help a robot arm follow a prescribed task-space trajectory~\cite{rakita2018relaxedik}, and to satisfy the kinematic constraints that arise when manipulating articulated objects~\cite{burget2013whole}.
Differential IK techniques can be used to follow task space trajectories, while satisfying constraints~\cite[\S 10.3]{siciliano2008springer}, \cite[\S 3.10]{tedrake2023manipulation}.

A key part of our work is the use of a smooth IK mapping to parametrize the constraint manifold.
Oftentimes, IK solutions are computed by solving a nonconvex mathematical program.
The tools of algebraic geometry can be used to reformulate certain IK problems as systems of polynomial equations, which can be solved as eigenvalue problems~\cite{raghavan1993inverse,nielsen1999kinematic,xie2022novel}.
However, neither of these methods yield closed-form IK solutions, nor do they guarantee smoothness.
Smooth IK solutions for certain 6DoF arms can be produced by dividing the joints into two sets of three, and treating each of these as ``virtual'' spherical joints~\cite[\S 2.12]{siciliano2009robotics}.
IK\nolinebreak{}Fast~\cite{diankov2010automated} can be used to automatically construct analytic IK solutions for broad classes of robot arm kinematics, and is available as part of the OpenRAVE toolkit~\cite{diankov2008openrave}.
Some arms have specifically-designed geometric solutions, such as the Universal Robotics UR-5/UR-10~\cite{hawkins2013analytic}.

Robot arms with more than six degrees of freedom have kinematic redundancy -- the arm can be moved while keeping its end effector fixed.
This is called \emph{self-motion} and is useful for avoiding obstacles and joint limits, but it implies that the forward kinematic mapping cannot be bijective.
\cite{hauser2020continuous} computes a globally-consistent pseudoinverse (discarding the redundancy), but this artificially restricts the configuration space.
Other approaches characterize the redundancy as a free parameter to be controlled in addition to the end-effector pose.
\cite{hemami1987more} presents a strategy for treating specific joints in a 7DoF arm as free parameters, reducing the problem to that of a 6DoF arm.
IKFast can discretize any additional joints.
Similar to the sphere-sphere 6DoF arms, certain 7DoF arms have a sphere-revolute-sphere kinematic structure (similar to the human arm), leading to elegant geometric solutions~\cite{hollerbach1985optimum,shimizu2008analytical}.
Specific geometric solutions are available for many common robot arms, including the KUKA iiwa~\cite{faria2018position}, Franka Emika Panda~\cite{he2021analytical}, and the Barrett WAM~\cite{singh2010analytical}.

Our parametrization can be combined with many planning algorithms to form a complete system.
In this paper, we specifically examine the canonical sampling-based planners: Rapidly-Exploring Random Trees (RRTs)~\cite{lavalle1998rapidly} and Probabilistic Roadmaps (PRMs)~\cite{kavraki1996probabilistic}.
Our contributions can also be used with the many extensions to these techniques~\cite{kuffner2000rrt,bohlin2000path,jaillet2010sampling,karaman2011sampling,ko2014randomized,salzman2016asymptotically,otte2016rrtx}.
We also describe how to use standard kinematic trajectory optimization techniques~\cite[\S 7.2]{tedrake2023manipulation}, \cite{zucker2013chomp,kalakrishnan2011stomp,toussaint2017tutorial}.
Finally, we describe how to extend the IRIS-NP algorithm~\cite{petersen2023growing} for computing convex collision-free sets to use our parametrization of the configuration space; such sets can be planned across with the GCS planning framework~\cite{marcucci2023motion}.
(These sets can also be used with other ``convex set planning algorithms''~\cite{marcucci2023fast,fernandez2018generative,deits2015efficient}.)

\section{Methodology}
\label{sec:methodology}

\begin{figure}
	\centering
	\begin{subfigure}[b]{0.49\linewidth}
		\centering
		\includegraphics[width=\linewidth]{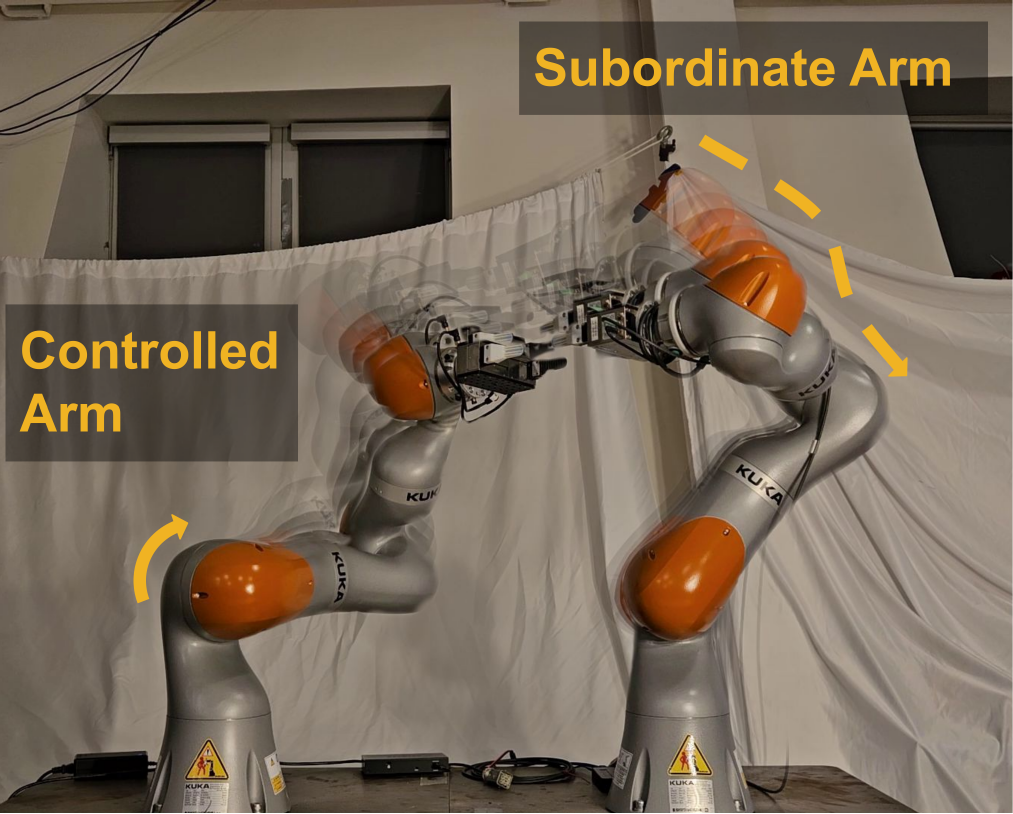}
	\end{subfigure}
	\hfill
	\begin{subfigure}[b]{0.49\linewidth}
		\centering
		\includegraphics[width=\linewidth]{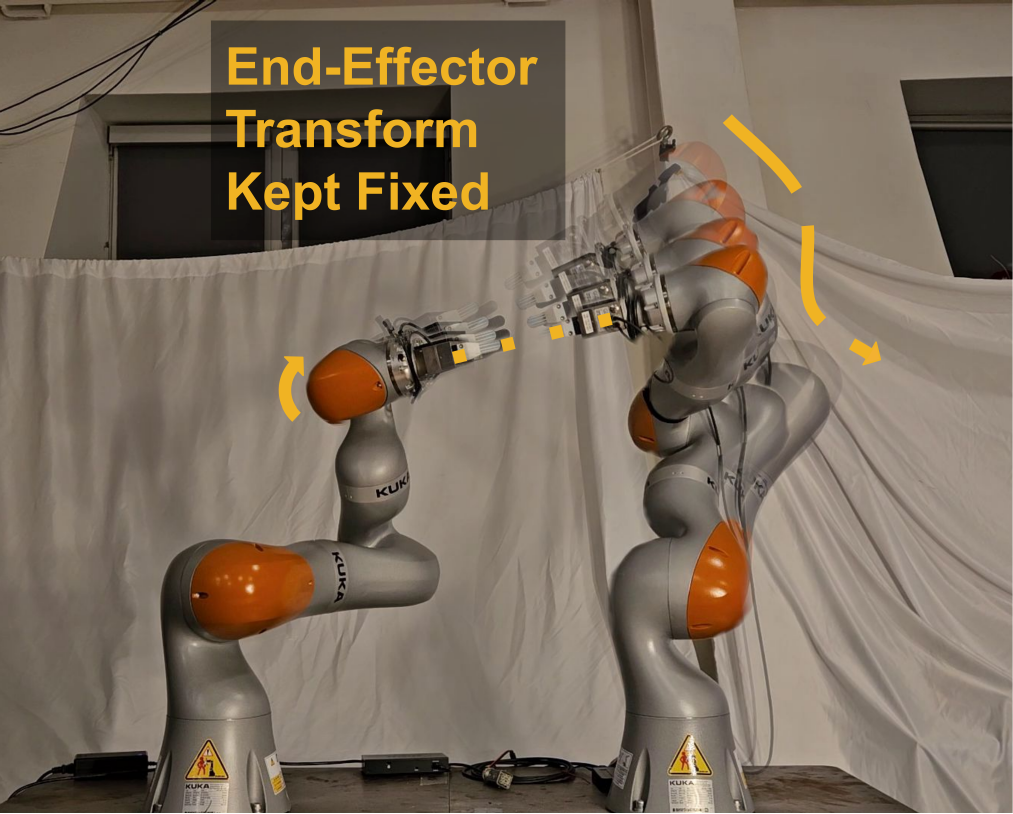}
	\end{subfigure}
	\caption{A high level description of our parametrization. The controlled arm can move freely, and analytic IK is used to position the subordinate arm so as to maintain a fixed transformation between the end-effectors.}
	\label{fig:parametrization}
\end{figure}

We introduce a bijective mapping between joint angles and end-effector pose for a single arm with analytic IK.
We then use this mapping to parametrize the set of valid configurations for constrained bimanual manipulation.
The joint angles of one arm are treated as free variables for the parametrized configuration space, and the aforementioned mapping is used to determine the joint angles for the other arms (visualized in \Cref{fig:parametrization}).
Finally, we explain the modifications needed to adapt existing planning algorithms to utilize this parametrization.

\subsection{Topology of Inverse Kinematics}
\label{sec:methodology:kinematic_topology}

The topological and geometric properties of inverse kinematic mappings are a classic area of study in robotics~\cite{gottlieb1988topology,burdick1989inverse,luck1993self}.
For an arm with $n\ge 6$ revolute joints, the configuration space is $\mc{C}\subseteq\mb{T}^n$, where $\mb{T}^n$ denotes the $n$-torus.
The forward kinematic mapping $f:\mc{C}\to\SE(3)$ computes the end-effector pose of the arm for a given choice of each joint angle.
We define the reachable set $\mc{X}=\set{f(\theta):\theta\in\mc{C}}\subseteq\SE(3)$.
To construct a homeomorphism between subsets of $\mc{C}$ and $\mc{X}$, we must restrict our domain of attention to avoid singular configurations, and augment $\mc{X}$ with additional degrees of freedom to match dimensions.

\begin{figure}
	\centering
	\begin{subfigure}[b]{0.49\linewidth}
		\centering
		\includegraphics[width=\linewidth]{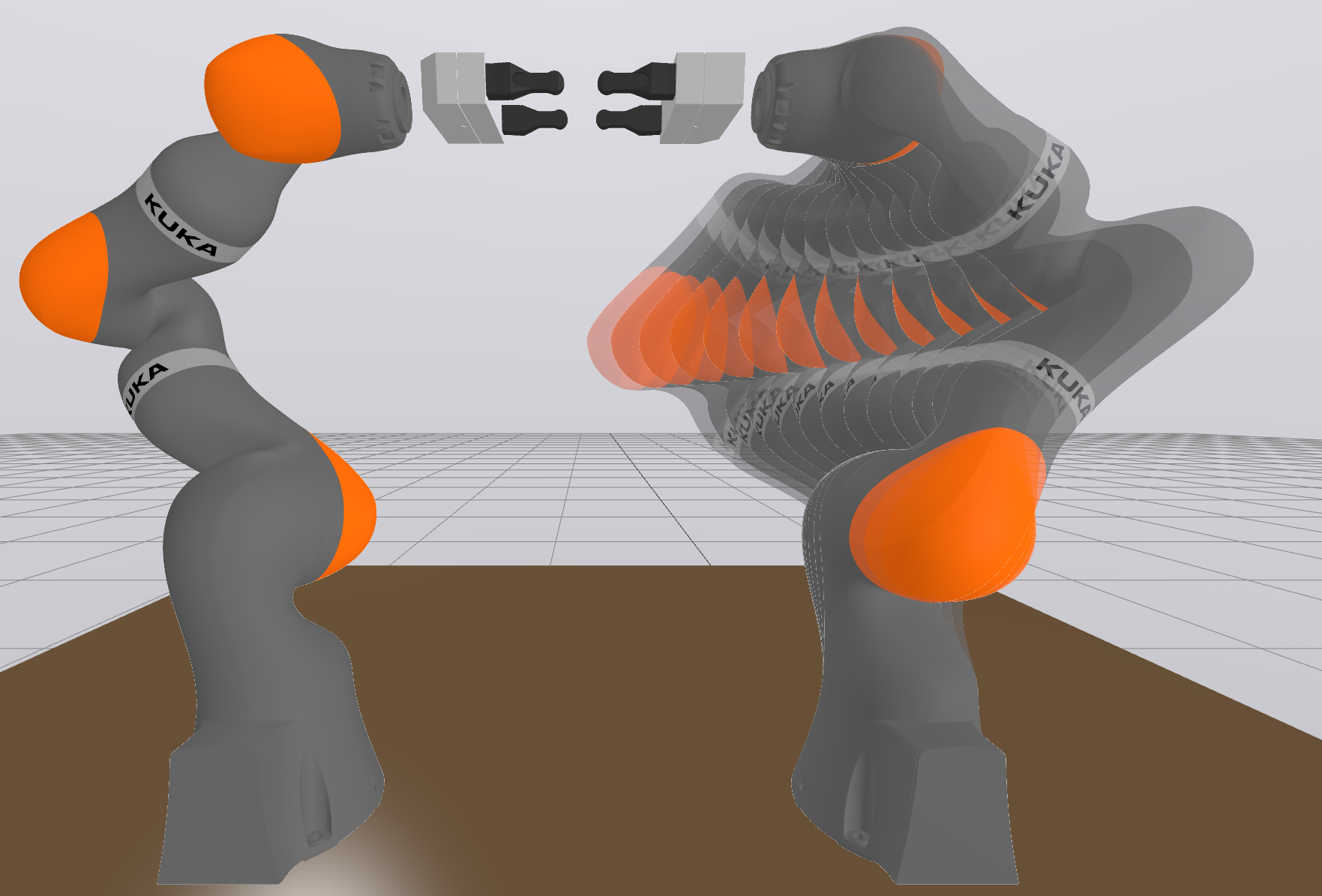}
	\end{subfigure}
	\hfill
	\begin{subfigure}[b]{0.49\linewidth}
		\centering
		\includegraphics[width=\linewidth]{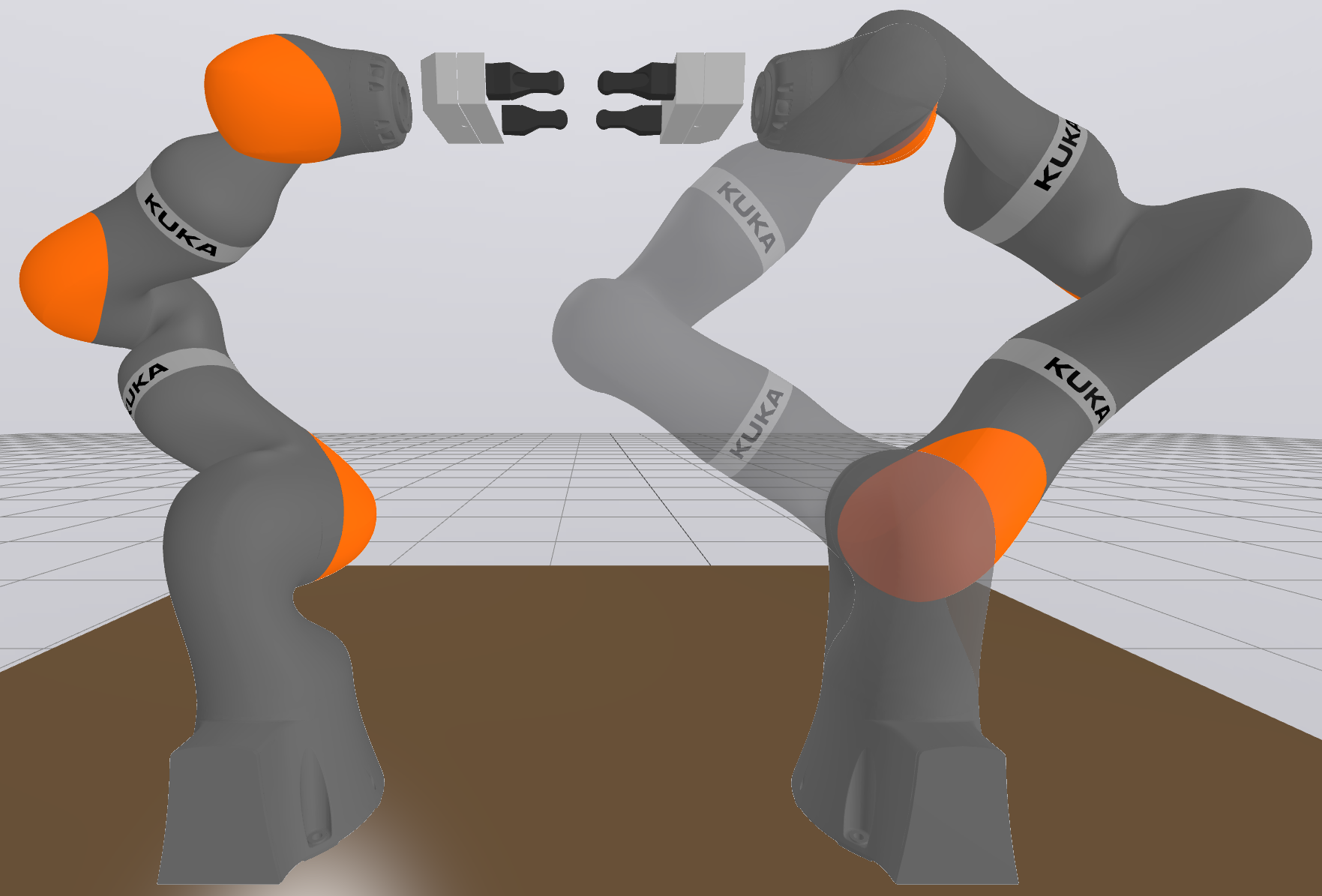}
	\end{subfigure}
	\caption{Continuous (left) and discrete (right) self-motions of a 7DoF arm. The continuous self-motion yields an additional degree of freedom for the planner to consider, whereas the discrete self-motion is not utilized.}
	\label{fig:self_motions}
\end{figure}

We give an overview of the terminology introduced in~\cite{burdick1989inverse} for describing the global behavior of inverse kinematic mappings.
A configuration for which the Jacobian of $f$ is full-rank is called a \emph{regular point}; otherwise, it is called a \emph{critical point}.
Because $f$ is not injective, the preimage of a single end-effector pose may contain only critical points, only regular points, or some of both; it is respectively called a \emph{critical value}, \emph{regular value}, and \emph{coregular value}.
\emph{$\mc{W}$-sheets} are the connected components of {regular values} in $\mc{X}$ whose boundaries are the coregular values of $f$.
The connected components of the preimages of $\mc{W}$-sheets are called \emph{$\mc{C}$-bundles} and are composed of regular points of $\mc{C}$.
For a regular value $x\in\mc{X}$, we have
\begin{equation}
	f\inv(x)=\bigcup_{i=1}^{m}\mc{M}_i(x),
	\label{eq:fk_preimage}
\end{equation}
where the $\mc{M}_i(x)$ are \emph{self-motion manifolds} of $x$, so called because motion within them does not affect the end-effector pose.
The label $i$ is called the \emph{global configuration parameter}, and a choice of $\psi\in\mc{M}_i(x)$ is called the \emph{redundancy parameter}. 
According to \cite{burdick1989inverse}, for robot arms in 3D space, the number of self-motion manifolds is at most 16; within a $\mc{C}$-bundle, the self-motion manifolds are homotopic; and if the arm has only revolute joints, then the self-motion manifolds are diffeomorphic to $\mb{T}^{n-6}$.
(If $n=6$, then the $\mc{M}_i$ are zero-dimensional, i.e., discrete points.)
Examples of the continuous and discrete self motions for a 7DoF arm are shown in \Cref{fig:self_motions}.

The $\mc{C}$-bundle/$\mc{W}$-sheet machinery allows us to construct well-defined IK mappings.
Let $\mc{W}_j\subseteq\mc{X}$ be a $\mc{W}$-sheet, and let $x_0\in\mc{W}_j$.
Then there is an smooth injection $g_{i,j}:\mc{W}_{j}\times\mc{M}_{i}(x_0)\to\mc{C}$.
Since the self-motion manifolds are homotopic within a $\mc{C}$-bundle, they are uniquely described in terms of their choice of $\mc{C}$-bundle and $\mc{W}$-sheet, so we use the shorthand $\mc{M}_{i,j}$ in place of $\mc{M}_i(x_0)$.
If we let $h_{i,j}$ map joint angles to their corresponding redundancy parameter, then $(f, h_{i,j})\of g_{i,j}$ is the identity mapping on $\mc{W}_{j}\times\mc{M}_{i,j}$.
Thus, with appropriate restrictions in domain and range, we have a bijection between the arm's joint angles and the product of its end-effector pose and redundancy parameters.
The set $\mc{C}_{i,j}$, defined as the image of $g_{i,j}$, is the set of joint angles which can be handled by these mappings.

\subsection{Parametrizing the Kinematically Constrained Space}
\label{sec:methodology:parametrization}

Now, we turn our attention to the bimanual case.
We use an additional subscript to denote which arm the sets and maps correspond to;
for example, $\mc{X}_{\leftarm}$ is the reachable set of the ``left'' arm, and $g_{i,j,\rightarm}$ denotes the inverse kinematic mapping for the ``right'' arm.

When a rigid object is held with both end effectors, a rigid transformation $\mc{T} \in \SE(3)$ between them becomes fixed; we let $\phi_{\mc{T}}:\mc{X}_\leftarm\to\SE(3)$ take in an end-effector pose for the left arm (henceforth called the \emph{controlled arm}), and output the target end-effector pose for the right arm (henceforth called the \emph{subordinate arm}).
We let $\mc{X}_{\mc{T}} := \{(x, \phi_{\mc{T}}(x)) : x \in \mc{X}_\leftarm\} \subset \mc{X}_\leftarm \times \SE(3)$ denote the space of end-effector poses which are feasible for the controlled arm and for which the pose of subordinate end-effector respects transformation $\mc{T}$.
Note that this latter pose may not be reachable for the subordinate arm, and a choice of redundancy parameter may require a violation of its joint limits.
We treat both of these cases as abstract obstacles in the configuration space.

For the remainder of the paper, we fix the global configuration parameter $i$ and choice of $\mc{W}$-sheet $j$ for the second arm.
Let $\mc{T}$ be the desired end-effector transformation.
We define a \emph{parametrized} configuration space $\mc{Q}:=\mc{C}_{\leftarm}\times\mc{M}_{i,j,\rightarm}$.
$q\in\mc{Q}$ determines joint angles for both arms via the mapping
\begin{equation}
	\xi:(\theta_\leftarm,\psi_\rightarm)\mapsto (\theta_\leftarm,g_{i,j,\rightarm}(\phi_{\mc{T}}(f_\leftarm(\theta_\leftarm)),\psi_\rightarm)).
	\label{eq:q_to_q_full}
\end{equation}
For more details on why we select this specific parametrization, see \Cref{sec:discussion}.
Let $\theta_{\min}$ and $\theta_{\max}$ be the lower and upper joint limits.
A configuration $(\theta_\leftarm,\psi_\rightarm)$ is valid if:
\begin{subequations}
	\label{eq:valid_conditions}
	\begin{align}
		\phi_{\mc{T}}(f_\leftarm(\theta_\leftarm))\in\mc{W}_{j,\rightarm} & \quad\ptxt{(Respect reachability.)} \label{eq:valid_conditions:require_reachable}\\
		\theta_{\min}\le\xi(\theta_\leftarm,\psi_\rightarm)\le\theta_{\max} & \quad\ptxt{(Respect joint limits.)} \label{eq:valid_conditions:require_respect_limits}
	\end{align}
\end{subequations}
We call the set of configurations satisfying these constraints $\mc{Q}_{\VALID}$.
For $q\in\mc{Q}$, if the robot is collision free for the joint angles $\xi(q)$, we say $q\in\mc{Q}_{\FREE}$.

\subsection{Reformulating the Motion Planning Problem}
\label{sec:methodology:problem_statement}

Let $\mf{s},\mf{t}\in\mc{C}_\leftarm\times\mc{C}_\rightarm$ be the start and goal configurations.
The \emph{constrained motion planning problem} requires finding a path $\gamma=(\gamma_\leftarm,\gamma_\rightarm):[0,1]\to\mc{C}_\leftarm\times\mc{C}_\rightarm$ by solving:
\begin{subequations}
	\newcommand{\alignspacing}{\;\;}
	\label{eq:constrained_planning}
	\begin{alignat}{3}
		\argmin & \alignspacing L(\gamma) \label{eq:constrained_planning:objective}\\
		\ptxt{s.t.} & \alignspacing \gamma(t) \ptxt{ collision free} & \alignspacing \forall t\in[0,1] \label{eq:constrained_planning:collision_free}\\
		& \alignspacing \phi_{\mc{T}}(f_\leftarm(\gamma_\leftarm(t)))=f_\rightarm(\gamma_\rightarm(t)) \; & \alignspacing \forall t\in[0,1] \label{eq:constrained_planning:kinematic_constraint}\\
		& \alignspacing \gamma(0)=\mf{s},\;\gamma(1)=\mf{t}. \label{eq:constrained_planning:start_goal}
	\end{alignat}
\end{subequations}
($L$ denotes the arc length functional, but can be replaced with another cost.)
The main challenge this formulation presents is the nonlinear equality constraint \eqref{eq:constrained_planning:kinematic_constraint}, as this requires $\gamma$ lie along a measure-zero set.
Trajectory optimizers may struggle with \eqref{eq:constrained_planning:kinematic_constraint}, and sampling-based planners must use one of the techniques described in \Cref{sec:related_work}.

Our \emph{parametrized motion planning problem} is written in terms of a trajectory $\bar\gamma:[0,1]\to\mc{Q}$, with start $\bar{\mf{s}}$ and goal $\bar{\mf{t}}$ satisfying $\xi(\bar{\mf{s}})=\mf{s}$ and $\xi(\bar{\mf{t}})=\mf{t}$:
\begin{subequations}
	\newcommand{\alignspacing}{\;\;}
	\label{eq:parametrized_problem}
	\begin{alignat}{3}
		\argmin & \alignspacing L(\xi\of\bar\gamma) \label{eq:parametrized_problem:objective}\\
		\ptxt{s.t.} & \alignspacing (\xi\of\bar\gamma)(t) \ptxt{ collision free} & \alignspacing \forall t\in[0,1] \label{eq:parametrized_problem:collision_free}\\
		& \alignspacing \bar\gamma(t)\in\mc{Q}_{\VALID} \; & \alignspacing \forall t\in[0,1] \label{eq:parametrized_problem:validity_constraint}\\
		& \alignspacing \bar\gamma(0)=\bar{\mf{s}},\;\bar\gamma(1)=\bar{\mf{t}}. \label{eq:parametrized_problem:start_goal}
	\end{alignat}
\end{subequations}
This formulation includes the implicit requirement that the entire planned trajectory be within a single $\mc{C}$-bundle, due to the restricted domain of $\xi$.
In \Cref{sec:results}, we demonstrate that this theoretical limitation is not a major roadblock to our framework's efficacy.
A major advantage of parametrization methods is that by construction, the end-effector poses $(f_\leftarm,f_\rightarm) \of \xi(\bar\gamma(t))$ are \emph{guaranteed} to be related by transformation $\mc{T}$.
For other methodologies, the constraints are only satisfied at discrete points along the trajectory.

\subsection{Motion Planning with the Parametrization}
\label{sec:methodology:motion_planning}

Constraint \eqref{eq:parametrized_problem:validity_constraint} is a nonlinear \emph{inequality} constraint, so feasible trajectories are constrained to lie in a positive volume set $\mc{Q}_{\VALID}\cap\mc{Q}_{\FREE}$.
Thus, unconstrained motion planning algorithms can function with only slight modifications.

\subsubsection{Sampling-Based Planning}
\label{sec:methodology:motion_planning:sampling_based}

The changes required for sampling based planners can be summarized as treating points outside $\mc{Q}_{\VALID}$ as being in collision.
Because $\mc{Q}_{\VALID}\cap\mc{Q}_{\FREE}$ has positive measure, rejection sampling can be used to draw valid samples.
When connecting samples (as in the ``Extend'' procedure of an RRT or ``Connect'' procedure of a PRM), the frequency with which collisions are checked must be adjusted, since distance in the parametrized space $\mc{Q}$ differs from distance in the full configuration space $\mc{C}_\leftarm\times\mc{C}_\rightarm$.
In particular, a small motion in $\mc{Q}$ can lead to a relatively large motion in $\mc{C}_\leftarm\times\mc{C}_\rightarm$, so collision checking must be done more frequently (or at a varying scale).

\subsubsection{Trajectory Optimization}
\label{sec:methodology:motion_planning:trajopt}

Trajectory optimization in configuration space is already nonconvex, so implementing constraints \eqref{eq:parametrized_problem:collision_free} and \eqref{eq:parametrized_problem:validity_constraint} requires no algorithmic changes.
As with sampling-based planning, collision avoidance (and other constraints applied to the full configuration space) must be enforced at a finer resolution.

\subsubsection{Graph of Convex Sets}
\label{sec:methodology:motion_planning:gcs}

Let $\mc{U}\subseteq\mc{Q}_{\VALID}\cap\mc{Q}_{\FREE}$ be convex.
Then the kinematic validity (and collision-free nature) of a linear path through $\mc{U}$ is guaranteed if its endpoints are contained in $\mc{U}$.
Thus, the Graph of Convex Sets Planner (GCS) can function as expected with two small modifications.
We minimize the arc length in the parametrized space $L(\bar\gamma)$, as this objective provides a useful convex surrogate for the true (nonconvex) objective \eqref{eq:parametrized_problem:objective}.
Also, for robot arms composed of revolute joints, the self-motion parameters are angle-valued, so one can either make cuts to the configuration space and treat it as Euclidean, or use the extension \emph{Graphs of Geodesically-Convex Sets} (GGCS)~\cite{cohn2023noneuclidean}.
The product of the angle-valued self-motion parameters will be a circle or n-torus, both of which admit a flat metric~\cite[p.345]{lee2012smooth}.
If we plan across geodesically convex (g-convex) subsets of $\mc{Q}_{\VALID}\cap\mc{Q}_{\FREE}$, then the problem satisfies the conditions presented in Assumptions 1 and 2 of~\cite{cohn2023noneuclidean}.
These assumptions guarantee that the resulting path will be kinematically valid and collision-free at all times.

\subsection{Constructing Convex Valid Sets}
\label{sec:methodology:constrained_iris}

To use (G)GCS, one must construct (g-)convex subsets of $\mc{Q}_{\VALID}\cap\mc{Q}_{\FREE}$.
The IRIS-NP algorithm~\cite{petersen2023growing} uses a counterexample search to find configurations in collision and constructs hyperplanes to separate the set from such configurations.
IRIS-NP can support inequality constraints beyond collision-avoidance, but they must be inequalities.
By running IRIS-NP through our parametrization, we avoid the equality constraints that would otherwise be present in our constrained bimanual manipulation problem.
Given a hyperellipsoid $\mc{E}(C,d)=\{q:\norm{q-d}_C^2\le 1\}$ (using the notation $\norm{q-d}_C^2=(q-d)\tpose C\tpose C(q-d)$), a halfspace intersection $\mc{H}(A,b)=\set{q:Aq\le b}$, and a \emph{constraint set} $\CS$, the \emph{generalized counterexample search program} is
\begin{subequations}
	\newcommand{\alignspacing}{\;\;}
	\label{eq:generalized_counterexample_search}
	\begin{alignat}{2}
		\min_q &\alignspacing \norm{q-d}_C^2 \label{eq:generalized_counterexample_search:objective}\\
		\ptxt{\st} &\alignspacing Aq\le b \label{eq:generalized_counterexample_search:halfspace}\\
		&\alignspacing q\not\in\CS. \label{eq:generalized_counterexample_search:constraint_violation}
	\end{alignat}
\end{subequations}

Given a bounding box $\mc{H}_0(A_0,b_0)$, a hyperellipsoid $\mc{E}(C,d)$ with $d\in\mc{H}_0(A_0,b_0)$, and a list of configuration-space constraints $\CS_1,\ldots,\CS_k$ to enforce, \Cref{alg:constrained_iris}  produces a halfspace intersection $\mc{H}(A,b)\subseteq\mc{H}_0(A_0,b_0)$ such that every point in $\mc{H}(A,b)$ satisfies the constraints.
\begin{algorithm}
	\caption{Constrained IRIS (Single Iteration)}
	\label{alg:constrained_iris}
	\begin{algorithmic}[1]
		\Require Bounding Box $\mc{H}_0(A_0,b_0)$\hfill\break
		Hyperellipsoid $\mc{E}(C,d)$ \st{} $d\in\mc{H}_0(A_0,b_0)$\hfill\break
		Constraint Sets $\CS_1,\ldots,\CS_k$
		\Ensure Halfspace Intersection $\mc{H}(A,b)$
		\State $A\gets A_0$, $b\gets b_0$
		\For{$\CS=\CS_1,\ldots,\CS_k$}
			\Repeat
				\State $(a^\star,b^\star)\gets\textsc{Solve}[\eqref{eq:generalized_counterexample_search}, \{A, b, C, d, \CS\}]$
				\State $A\gets\textsc{VStack}(A, a^\star),b\gets\textsc{VStack}(b,b^\star)$
			\Until $\textsc{Infeasible}$
		\EndFor
		\State\Return $\mc{H}(A,b)$
	\end{algorithmic}
\end{algorithm}

We now describe the {constraint sets} $\CS$ needed for \Cref{alg:constrained_iris} to generate g-convex sets in $\mc{Q}_{\VALID}\cap\mc{Q}_{\FREE}$, and how to encode \eqref{eq:generalized_counterexample_search:constraint_violation}.
For $q=(\theta_\leftarm,\psi_\rightarm)$ (or $q=(\theta_\leftarm,\psi_\rightarm,\mc{T})$ if the end effector transformation is allowed to vary), consider the auxiliary variable $\theta_\rightarm$ denoting the joint angles of the subordinate arm, computed with $(\theta_\leftarm,\theta_\rightarm)=\xi(q)$.

First, we require that any inverse trigonometric functions used in the analytic IK mapping $g_{i,j,\rightarm}$ do not violate their domains.
Although this constraint would be enforced by the later constraints, specifically handling this case first greatly improves the performance of the later counterexample searches.
For example, \cite[Eq. 4]{faria2018position} takes the $\arccos$ of an argument $w$, so we encode \eqref{eq:generalized_counterexample_search:constraint_violation} as $\abs{w}\ge 1+\epsilon$.
When using the analytic IK solution for the KUKA iiwa, we enforce this constraint for equations (4), (6), (18), and (23) of \cite{faria2018position}.

Next, we check the joint limits \eqref{eq:valid_conditions:require_respect_limits}, encoded for \eqref{eq:generalized_counterexample_search:constraint_violation} as
\begin{equation*}
	\max(\xi(q)-\theta_{\max},\theta_{\min}-\xi(q))\ge\epsilon.
\end{equation*}

Finally, a configuration $q$ is said to be \emph{reachable} if $\phi_{\mc T}(f_\leftarm(\theta_\leftarm))=f_\rightarm(\theta_\rightarm)$.
Although this is an equality constraint, the set of configurations satisfying the constraint has positive volume in the parametrized space, so \Cref{alg:constrained_iris} can still be used to generate a convex inner-approximation.
For reachability counterexamples \eqref{eq:valid_conditions:require_reachable}, we compute the squared Frobenius norm of the difference between desired and realized end-effector pose, encoding \eqref{eq:generalized_counterexample_search:constraint_violation} as
\begin{equation*}
	\norm{\phi_{\mc{T}}(f_\leftarm(\theta_\leftarm))-f_\rightarm(\theta_\rightarm)}_F^2\ge\epsilon.
\end{equation*}

These three constraints will ensure $\mc{H}(A,b)\subseteq\mc{Q}_{\VALID}$.
To also enforce $\mc{H}(A,b)\subseteq\mc{Q}_{\FREE}$, we search for configurations $q$ such that the robot is in collision.
We separately find counterexamples for each pair of collision bodies, using equation (2) of \cite{petersen2023growing}.
Note that this equation operates on the full configuration $(\theta_\leftarm,\theta_\rightarm)$, as obtained from the parametrized configuration with $\xi$.
Because \eqref{eq:generalized_counterexample_search} is a nonlinear program, we solve it using SNOPT~\cite{snopt} with random initializations until a solution is obtained or a predefined number of consecutive failures is reached (and in that case, return infeasible).

\section{Results}
\label{sec:results}

We demonstrate our new constrained planning framework using a bimanual manipulation setup with two \emph{KUKA iiwa} 7DoF arms.
Interactive recordings of all trajectories are available online at \href{https://cohnt.github.io/Bimanual-Web/index.html}{https://cohnt.github.io/Bimanual-Web/}.
We use the analytic IK map presented in~\cite{faria2018position}.
To evaluate the merits of our IK parametrization for constrained planning, we consider a task where the two arms must move an object around a set of shelves, while avoiding collisions.
We test four approaches under our parametrization:
\begin{itemize}
	\item[1a.] \emph{IK-BiRRT.} We use the single-query bidirectional RRT (BiRRT) algorithm~\cite{kuffner2000rrt}.
	\item[2a.] \emph{IK-Trajopt} We directly solve \eqref{eq:parametrized_problem} with kinematic trajectory optimization~\cite[\S 10.3]{tedrake2023underactuated}, using the Drake modeling toolbox~\cite{tedrake2019drake}.
	We use the output of the BiRRT planner as the initial guess for the trajectory optimizer.
	\item[3a.] \emph{IK-PRM.} We use the multi-query PRM algorithm~\cite{kavraki1996probabilistic}, initialized with nodes from multiple BiRRTs to ensure connectivity, as in~\cite[\S C]{marcucci2023motion}.
	\item[4a.]\emph{IK-GCS}. We use GCS-planner~\cite{marcucci2023motion} with $19$ regions, constructed from hand-selected seed points.
\end{itemize}
For both the BiRRT and PRM plans, we use short-cutting to post-process the paths~\cite{schwarzer2004exact}.
We solve the GCS problems with Mosek~\cite{mosek}.
We compare these parametrized planners with constrained planning baselines.
\begin{itemize}
	\item[1b.] \emph{Constrained Trajectory Optimization.} We solve \eqref{eq:constrained_planning} with kinematic trajectory optimization, using the IK-BiRRT plan as the initial guess to compare with IK-Trajopt.
	\item[2b.] \emph{Sampling-Based Planning.} For sampling-based planners, we use the single-query Atlas-BiRRT and multi-query Atlas-PRM algorithms~\cite{kingston2019exploring}, as implemented in the Open Motion Planning Library~\cite{sucan2012ompl}.
	The atlas and PRM are initialized from multiple Atlas-BiRRT runs.
\end{itemize}
We do not compare to any GCS baseline without IK, as the constraint manifold is inherently nonconvex; IK-GCS is the first proposal for extending GCS to this class of problems.

\begin{table}
	\centering
	\renewcommand{\arraystretch}{1.05}
	\begin{tabular}{|c|c|c|c|} \hline
		Method & Top to Middle & Middle to Bottom & Bottom to Top\\ \hline
		Trajopt & 4.58* & 2.85* & \textbf{4.35*} \\ \hline
		Atlas-BiRRT & 4.72 & 5.04 & 6.61 \\ \hline
		Atlas-PRM & 5.43 & 5.67 & 6.99 \\ \hline
		IK-Trajopt & 4.24* & \textbf{1.81*} & 8.87 \\ \hline
		IK-BiRRT & 9.91 & 8.69 & 11.42 \\ \hline
		IK-PRM & 4.67 & 8.93 & 9.21 \\ \hline
		IK-GCS & \textbf{2.09} & 3.32 & 5.62 \\ \hline
	\end{tabular}
	\caption{
		Path lengths (measured in configuration space) for each method with various start and goal configurations.
		Paths marked with an asterisk were not collision-free.
	}
	\label{tab:path_length}
\end{table}

\begin{table}
	\centering
	\renewcommand{\arraystretch}{1.05}
	\begin{tabular}{|c|c|c|c|} \hline
		Method & Top to Middle & Middle to Bottom & Bottom to Top\\ \hline
		Trajopt & 10.37 & 5.36 & 7.25 \\ \hline
		Atlas-BiRRT & 140.82 & 155.91 & 201.32 \\ \hline
		Atlas-PRM & 0.69 & 0.86 & 0.96 \\ \hline
		IK-Trajopt & 19.48 & 18.70 & 22.29 \\ \hline
		IK-BiRRT & 49.42 & 52.53 & 54.10 \\ \hline
		IK-PRM & \textbf{0.46} & \textbf{0.64} & \textbf{0.61} \\ \hline
		IK-GCS & 3.41 & 2.32 & 3.32 \\ \hline
	\end{tabular}
	\caption{
		Online planning time (in seconds) for each method with various start and goal configurations.
		Atlas-BiRRT runtimes were only averaged over successful runs (not including timeouts).
	}
	\label{tab:online_runtime}
\end{table}

\begin{figure*}[t]
	\centering
	\begin{subfigure}[b]{0.26\linewidth}
		\centering
		\includegraphics[width=\linewidth]{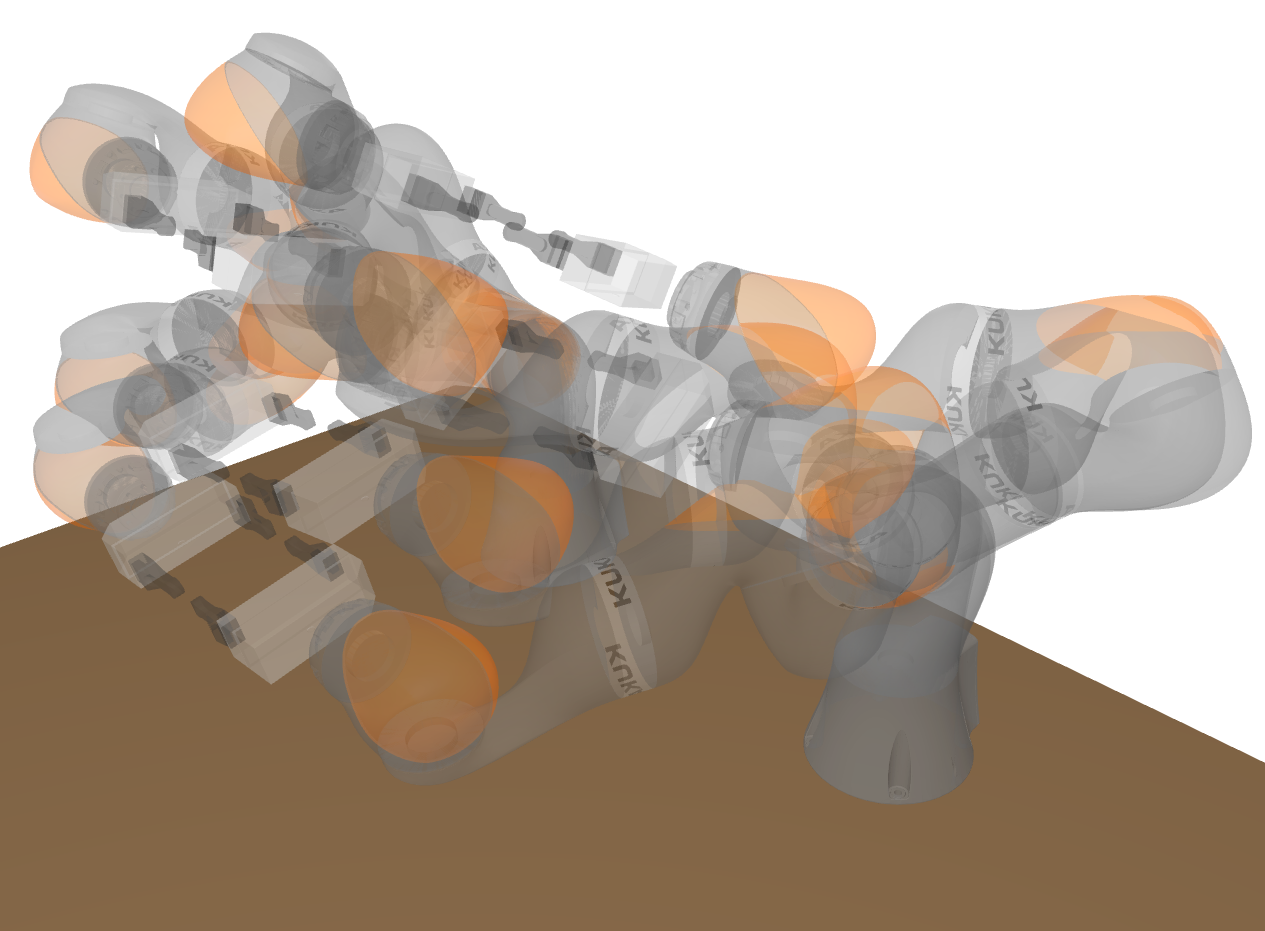}
		\caption{The coverage of a single convex region, as presented by a random collection of configurations.}
		\label{fig:iris:free_space}
	\end{subfigure}
	\hfill
	\begin{subfigure}[b]{0.33\linewidth}
		\centering
		\includegraphics[width=\linewidth]{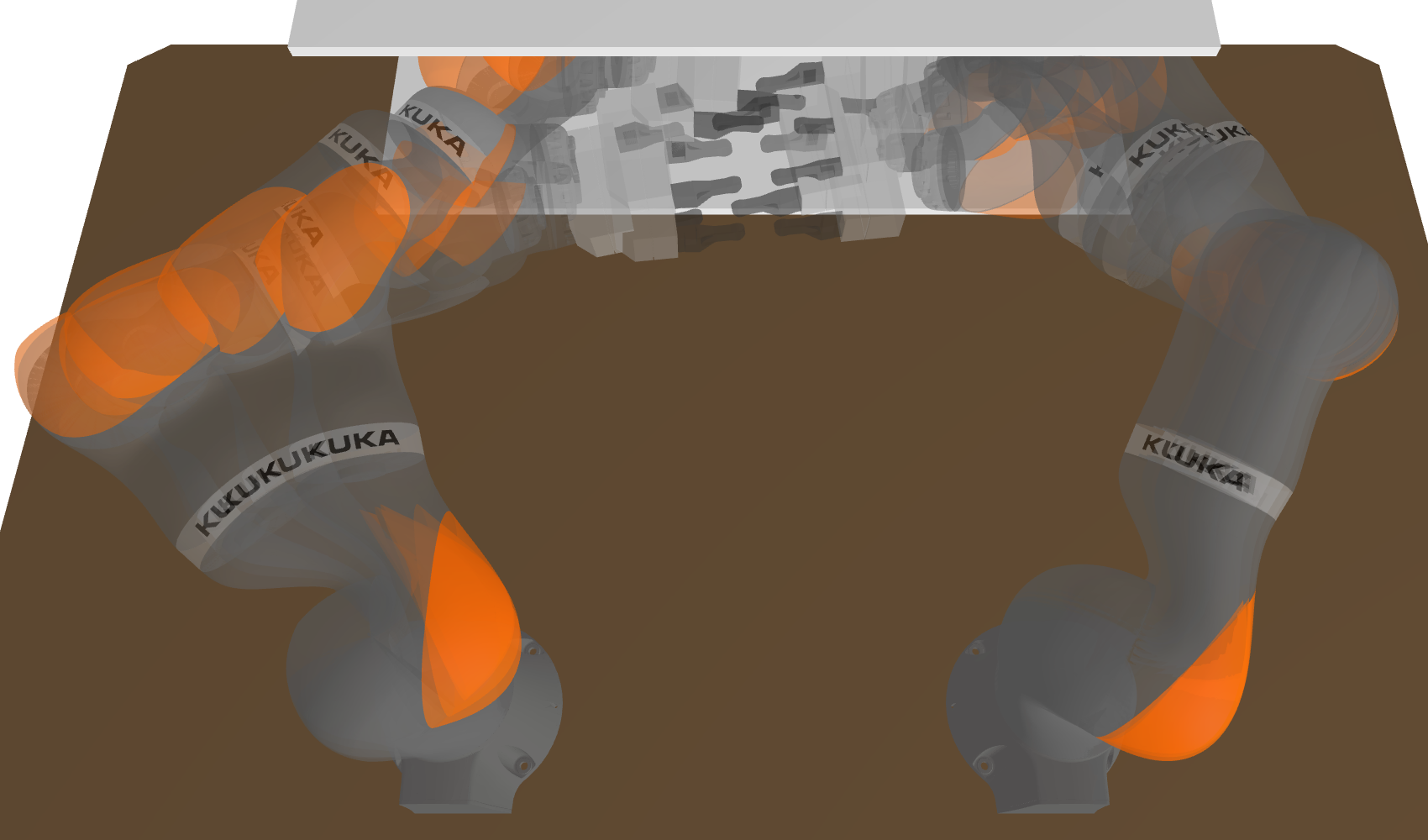}
		\caption{A collection of configurations from a region that represents motions that reach into and out of a shelf.}
		\label{fig:iris:shelves}
	\end{subfigure}
	\hfill
	\begin{subfigure}[b]{0.33\linewidth}
		\centering
		\includegraphics[width=\linewidth]{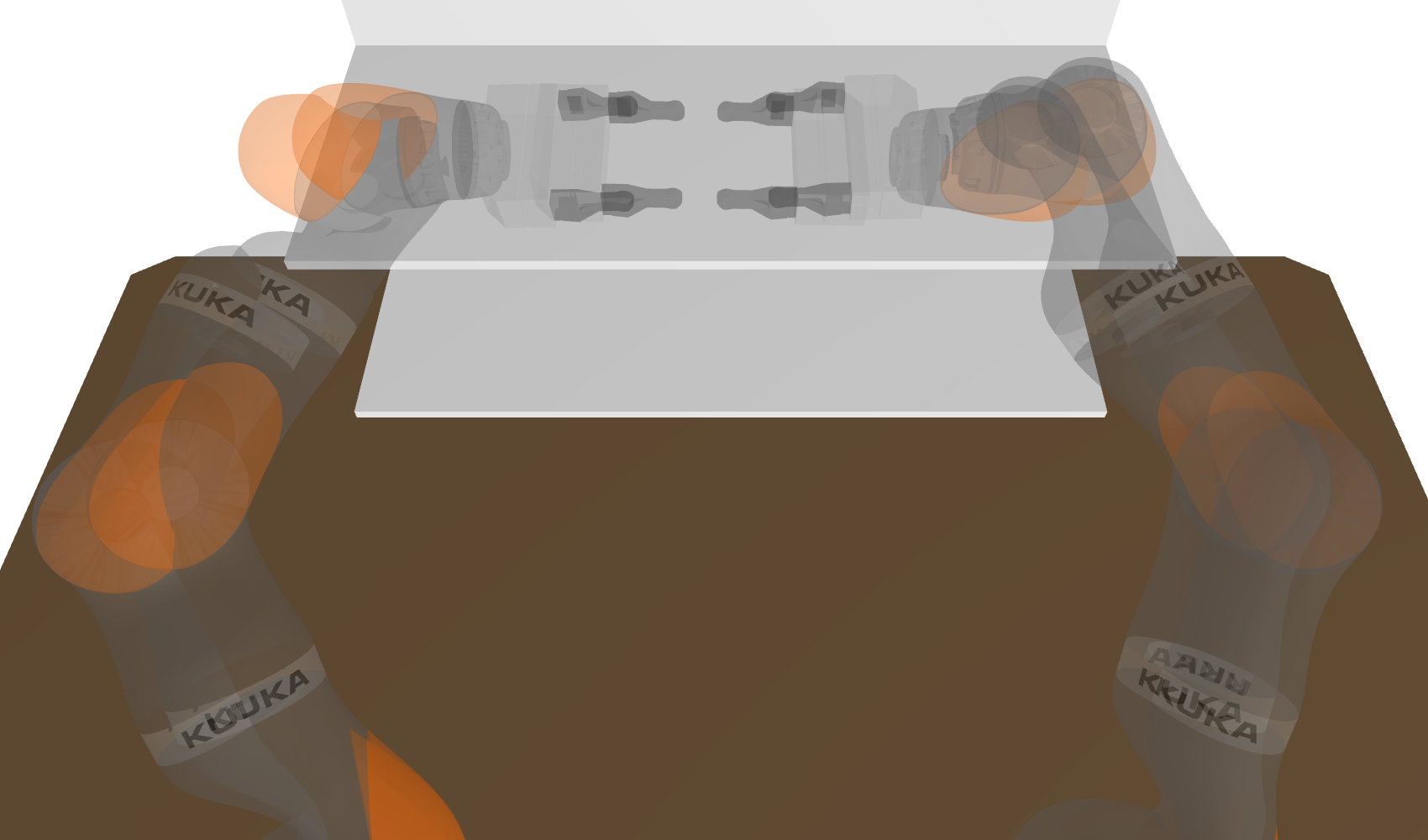}
		\caption{A region that represents varying grasp distances, in addition to collision-free configurations in the shelf (not shown).}
		\label{fig:iris:shelves_vary_grasp}
	\end{subfigure}
	\caption{
		Robot configurations sampled from various IRIS regions.
	}
	\label{fig:iris}
\end{figure*}

\textbf{Constraint Violations:}
Because the baseline methods can only enforce the kinematic constraint at discrete points, the constraint violation can be significant between such points.
The OMPL planners experienced a maximum constraint violation of 6.62 cm, and the trajectory optimization baseline experienced a maximum constraint violation of 3.22 cm.
In comparison, our parametrization methods maintained all constraints within 0.001 cm.
Plans from the trajectory optimization baseline also had slight collisions with obstacles.

\textbf{Path Length \& Planning Time:}
Across all methods, for various start and goal configurations, we compare path length in \Cref{tab:path_length} and online planning time in \Cref{tab:online_runtime}.
We ran both BiRRT methods 10 times for each plan, and report the average path length and planning time.
We set a maximum planning time of 10 minutes for Atlas-BiRRT, and omit these from the averaging.
Out of the 30 runs used for the table, Atlas-BiRRT timed out twice.
IK-BiRRT never timed out; the longest plan took 81.17 seconds to compute.

\Cref{tab:online_runtime} does not include offline compute time.
The time to construct the Atlas-PRM varies greatly; with three random seeds, it took 33.25, 437.93, and 554.90 seconds.
Constructing the IK-PRM took 2648.79 seconds, and constructing the IRIS regions for GCS took 18361.36 seconds (966.39 seconds per region on average).
The IRIS region construction can also be parallelized, improving runtime.

Overall, the PRM methods have the shortest online runtimes.
GCS is consistently faster than the other optimization approaches, as a result of the offline precomputation of IRIS regions.
The other optimization approaches are sometimes able to find shorter paths than GCS, since they have fewer constraints, but can get stuck in local minima. %
Although the atlas methodologies may find shorter paths than their IK counterparts, this is at the cost of significantly higher runtimes and potentially large kinematic constraint violations.

\textbf{Task Space Coverage of IRIS Regions.}
In \Cref{fig:iris}, we superimpose the end-effector poses from many sampled bimanual configurations within individual IRIS regions.
Despite the complicated nonlinear mapping, these convex sets are able to cover large swaths of task space, as shown in \Cref{fig:iris:free_space}.
In \Cref{fig:iris:shelves}, we demonstrate that IRIS regions can reliably encompass the motions required to reach into and out of a shelf.
And in \Cref{fig:iris:shelves_vary_grasp}, we visualize an IRIS region that allows the grasp distance to vary.
This is accomplished by treating the grasp distance as an additional degree of freedom in \Cref{alg:constrained_iris}.
GCS can use such regions to plan motions for objects of different sizes; we include hardware demonstrations in our results video.

\section{Discussion}
\label{sec:discussion}

We presented a novel parametrization of the constrained configuration space that arises in bimanual manipulation, which can be leveraged by both sampling-based planners and trajectory optimizers for more efficient planning.
Our parametrization can be used to find shorter paths more quickly than existing approaches, and these paths will satisfy the kinematic constraints at all points along the trajectory.
This parametrization also enables the use of planners such as GCS, which previously could not be applied to configuration spaces with nonlinear equality constraints.

Other parametrizations for the constrained configuration space are symmetric, and may seem more natural:
\begin{enumerate}
	\item Treating the end-effector configuration and redundancy parameters for both arms as the free variables, and using analytic IK for both arms.
	\item Treating the first four joints of each arm as free variables, and solving IK for the remaining six joints as a virtual 6DoF arm whose middle link is represented by the object held by both end-effectors.
\end{enumerate}
But these choices present other disadvantages.

For the first option, we would have to choose global configuration parameters for both arms; in the case of the KUKA iiwa, this involves 64 choices (instead of the 8 options for our parametrization).
Also, the shortest paths for the end effector may lead to very inefficient paths in joint space -- our parametrization can at least minimize the joint space distance for one arm.
Finally, it requires planning over $\operatorname{SO}(3)$, which is not possible for GCS (see~\cite[Thm. 5]{cohn2023noneuclidean}).

For the second option, the choice of end-effector transformation $\mc{T}$ determines the kinematic structure of the virtual arm, so different grasps would require different analytic IK solutions.
Constructing such solutions would be time-consuming, and they may not always exist.

There are clear directions for future work.
Enabling the planner to move between $\mc C$-bundles would unlock a greater variety of arm motions, potentially allowing the selected planner to compute shorter paths.
Another important future step is to explicitly consider singular configurations.
The IK mapping used in this work is still defined at singular configurations, but nearby configurations may violate the subordinate arm joints limits or reachability constraints.
This makes the IRIS region generation process implicitly avoid singular configurations, but the other planning methodologies may not detect infeasible configurations near singularities.

\section*{Acknowledgement}
\label{sec:acknowledgement}

The authors thank Zak Kingston (Rice University) and Josep Porta (Polytechnic University of Catalonia) for their comments and suggestions, and for advice on the implementation of the baselines.

\bibliographystyle{IEEEtran}
\bibliography{IEEEabrv,ref}

\begin{thebibliography}{10}
\providecommand{\url}[1]{#1}
\csname url@rmstyle\endcsname
\providecommand{\newblock}{\relax}
\providecommand{\bibinfo}[2]{#2}
\providecommand\BIBentrySTDinterwordspacing{\spaceskip=0pt\relax}
\providecommand\BIBentryALTinterwordstretchfactor{4}
\providecommand\BIBentryALTinterwordspacing{\spaceskip=\fontdimen2\font plus
\BIBentryALTinterwordstretchfactor\fontdimen3\font minus
  \fontdimen4\font\relax}
\providecommand\BIBforeignlanguage[2]{{%
\expandafter\ifx\csname l@#1\endcsname\relax
\typeout{** WARNING: IEEEtran.bst: No hyphenation pattern has been}%
\typeout{** loaded for the language `#1'. Using the pattern for}%
\typeout{** the default language instead.}%
\else
\language=\csname l@#1\endcsname
\fi
#2}}

\bibitem{krebs2022bimanual}
F.~Krebs and T.~Asfour, ``A bimanual manipulation taxonomy,'' \emph{IEEE
  Robotics and Automation Letters}, vol.~7, no.~4, pp. 11\,031--11\,038, 2022.

\bibitem{holladay2024robust}
\BIBentryALTinterwordspacing
R.~Holladay, T.~Lozano-Pérez, and A.~Rodriguez, ``Robust planning for
  multi-stage forceful manipulation,'' \emph{The International Journal of
  Robotics Research}, vol.~43, no.~3, pp. 330--353, 2024. [Online]. Available:
  \url{https://doi.org/10.1177/02783649231198560}
\BIBentrySTDinterwordspacing

\bibitem{berenson2009manipulation}
D.~Berenson, S.~S. Srinivasa, D.~Ferguson, and J.~J. Kuffner, ``Manipulation
  planning on constraint manifolds,'' in \emph{2009 IEEE international
  conference on robotics and automation}.\hskip 1em plus 0.5em minus
  0.4em\relax IEEE, 2009, pp. 625--632.

\bibitem{krauskopf2007numerical}
B.~Krauskopf, H.~M. Osinga, and J.~Gal{\'a}n-Vioque, \emph{Numerical
  continuation methods for dynamical systems}.\hskip 1em plus 0.5em minus
  0.4em\relax Springer, 2007, vol.~2.

\bibitem{bonilla2015sample}
M.~Bonilla, E.~Farnioli, L.~Pallottino, and A.~Bicchi, ``Sample-based motion
  planning for soft robot manipulators under task constraints,'' in \emph{2015
  IEEE International Conference on Robotics and Automation (ICRA)}.\hskip 1em
  plus 0.5em minus 0.4em\relax IEEE, 2015, pp. 2522--2527.

\bibitem{pavone2019trajectory}
R.~Bonalli, A.~Cauligi, A.~Bylard, T.~Lew, and M.~Pavone, ``Trajectory
  optimization on manifolds: A theoretically-guaranteed embedded sequential
  convex programming approach,'' in \emph{Proceedings of Robotics: Science and
  Systems}, Freiburgim Breisgau, Germany, June 2019.

\bibitem{marcucci2024shortest}
T.~Marcucci, J.~Umenberger, P.~Parrilo, and R.~Tedrake, ``Shortest paths in
  graphs of convex sets,'' \emph{SIAM Journal on Optimization}, vol.~34, no.~1,
  pp. 507--532, 2024.

\bibitem{marcucci2023motion}
T.~Marcucci, M.~Petersen, D.~von Wrangel, and R.~Tedrake, ``Motion planning
  around obstacles with convex optimization,'' \emph{Science robotics}, vol.~8,
  no.~84, p. eadf7843, 2023.

\bibitem{bonilla2017noninteracting}
M.~Bonilla, L.~Pallottino, and A.~Bicchi, ``Noninteracting constrained motion
  planning and control for robot manipulators,'' in \emph{2017 IEEE
  International Conference on Robotics and Automation (ICRA)}.\hskip 1em plus
  0.5em minus 0.4em\relax IEEE, 2017, pp. 4038--4043.

\bibitem{stilman2010global}
M.~Stilman, ``Global manipulation planning in robot joint space with task
  constraints,'' \emph{IEEE Transactions on Robotics}, vol.~26, no.~3, pp.
  576--584, 2010.

\bibitem{mirabel2016hpp}
J.~Mirabel, S.~Tonneau, P.~Fernbach, A.-K. Sepp{\"a}l{\"a}, M.~Campana,
  N.~Mansard, and F.~Lamiraux, ``Hpp: A new software for constrained motion
  planning,'' in \emph{2016 IEEE/RSJ International Conference on Intelligent
  Robots and Systems (IROS)}.\hskip 1em plus 0.5em minus 0.4em\relax IEEE,
  2016, pp. 383--389.

\bibitem{jaillet2012path}
L.~Jaillet and J.~M. Porta, ``Path planning under kinematic constraints by
  rapidly exploring manifolds,'' \emph{IEEE Transactions on Robotics}, vol.~29,
  no.~1, pp. 105--117, 2012.

\bibitem{bohigas2013planning}
O.~Bohigas, M.~E. Henderson, L.~Ros, M.~Manubens, and J.~M. Porta, ``Planning
  singularity-free paths on closed-chain manipulators,'' \emph{IEEE
  Transactions on Robotics}, vol.~29, no.~4, pp. 888--898, 2013.

\bibitem{kim2016tangent}
B.~Kim, T.~T. Um, C.~Suh, and F.~C. Park, ``Tangent bundle rrt: A randomized
  algorithm for constrained motion planning,'' \emph{Robotica}, vol.~34, no.~1,
  pp. 202--225, 2016.

\bibitem{kingston2019exploring}
Z.~Kingston, M.~Moll, and L.~E. Kavraki, ``Exploring implicit spaces for
  constrained sampling-based planning,'' \emph{The International Journal of
  Robotics Research}, vol.~38, no. 10-11, pp. 1151--1178, 2019.

\bibitem{han2008convexly}
L.~Han, L.~Rudolph, J.~Blumenthal, and I.~Valodzin, ``Convexly stratified
  deformation spaces and efficient path planning for planar closed chains with
  revolute joints,'' \emph{The International Journal of Robotics Research},
  vol.~27, no. 11-12, pp. 1189--1212, 2008.

\bibitem{mcmahon2018sampling}
T.~McMahon, S.~Thomas, and N.~M. Amato, ``Sampling-based motion planning with
  reachable volumes for high-degree-of-freedom manipulators,'' \emph{The
  International Journal of Robotics Research}, vol.~37, no.~7, pp. 779--817,
  2018.

\bibitem{zha2018learning}
F.~Zha, Y.~Liu, W.~Guo, P.~Wang, M.~Li, X.~Wang, and J.~Li, ``Learning the
  metric of task constraint manifolds for constrained motion planning,''
  \emph{Electronics}, vol.~7, no.~12, p. 395, 2018.

\bibitem{csucan2012motion}
I.~A. {\c{S}}ucan and S.~Chitta, ``Motion planning with constraints using
  configuration space approximations,'' in \emph{2012 IEEE/RSJ International
  Conference on Intelligent Robots and Systems}.\hskip 1em plus 0.5em minus
  0.4em\relax IEEE, 2012, pp. 1904--1910.

\bibitem{kingston2018sampling}
Z.~Kingston, M.~Moll, and L.~E. Kavraki, ``Sampling-based methods for motion
  planning with constraints,'' \emph{Annual review of control, robotics, and
  autonomous systems}, vol.~1, pp. 159--185, 2018.

\bibitem{bordalba2022direct}
R.~Bordalba, T.~Schoels, L.~Ros, J.~M. Porta, and M.~Diehl, ``Direct
  collocation methods for trajectory optimization in constrained robotic
  systems,'' \emph{IEEE Transactions on Robotics}, 2022.

\bibitem{han2001kinematics}
L.~Han and N.~M. Amato, ``A kinematics-based probabilistic roadmap method for
  closed chain systems,'' in \emph{Algorithmic and computational
  robotics}.\hskip 1em plus 0.5em minus 0.4em\relax AK Peters/CRC Press, 2001,
  pp. 243--251.

\bibitem{cortes2005sampling}
J.~Cort{\'e}s and T.~Sim{\'e}on, ``Sampling-based motion planning under
  kinematic loop-closure constraints,'' in \emph{Algorithmic Foundations of
  Robotics VI}.\hskip 1em plus 0.5em minus 0.4em\relax Springer, 2005, pp.
  75--90.

\bibitem{gharbi2008sampling}
M.~Gharbi, J.~Cort{\'e}s, and T.~Simeon, ``A sampling-based path planner for
  dual-arm manipulation,'' in \emph{2008 IEEE/ASME International Conference on
  Advanced Intelligent Mechatronics}.\hskip 1em plus 0.5em minus 0.4em\relax
  IEEE, 2008, pp. 383--388.

\bibitem{wang2019inverse}
J.~Wang, S.~Liu, B.~Zhang, and C.~Yu, ``Inverse kinematics-based motion
  planning for dual-arm robot with orientation constraints,''
  \emph{International Journal of Advanced Robotic Systems}, vol.~16, no.~2, p.
  1729881419836858, 2019.

\bibitem{kanehiro2012efficient}
F.~Kanehiro, E.~Yoshida, and K.~Yokoi, ``Efficient reaching motion planning and
  execution for exploration by humanoid robots,'' in \emph{2012 IEEE/RSJ
  International Conference on Intelligent Robots and Systems}.\hskip 1em plus
  0.5em minus 0.4em\relax IEEE, 2012, pp. 1911--1916.

\bibitem{rakita2018relaxedik}
D.~Rakita, B.~Mutlu, and M.~Gleicher, ``Relaxedik: Real-time synthesis of
  accurate and feasible robot arm motion.'' in \emph{Robotics: Science and
  Systems}, vol.~14.\hskip 1em plus 0.5em minus 0.4em\relax Pittsburgh, PA,
  2018, pp. 26--30.

\bibitem{burget2013whole}
F.~Burget, A.~Hornung, and M.~Bennewitz, ``Whole-body motion planning for
  manipulation of articulated objects,'' in \emph{2013 IEEE International
  Conference on Robotics and Automation}.\hskip 1em plus 0.5em minus
  0.4em\relax IEEE, 2013, pp. 1656--1662.

\bibitem{siciliano2008springer}
B.~Siciliano, O.~Khatib, and T.~Kr{\"o}ger, \emph{Springer handbook of
  robotics}.\hskip 1em plus 0.5em minus 0.4em\relax Springer, 2008, vol. 200.

\bibitem{tedrake2023manipulation}
\BIBentryALTinterwordspacing
R.~Tedrake, \emph{Robotic Manipulation}, 2023. [Online]. Available:
  \url{http://manipulation.mit.edu}
\BIBentrySTDinterwordspacing

\bibitem{raghavan1993inverse}
\BIBentryALTinterwordspacing
M.~Raghavan and B.~Roth, ``Inverse kinematics of the general 6r manipulator and
  related linkages,'' \emph{Journal of Mechanical Design}, vol. 115, no.~3, pp.
  502--508, sep 1993. [Online]. Available:
  \url{https://doi.org/10.1115/1.2919218}
\BIBentrySTDinterwordspacing

\bibitem{nielsen1999kinematic}
J.~Nielsen and B.~Roth, ``On the kinematic analysis of robotic mechanisms,''
  \emph{The International Journal of Robotics Research}, vol.~18, no.~12, pp.
  1147--1160, 1999.

\bibitem{xie2022novel}
S.~Xie, L.~Sun, G.~Chen, Z.~Wang, and Z.~Wang, ``A novel solution to the
  inverse kinematics problem of general 7r robots,'' \emph{IEEE Access},
  vol.~10, pp. 67\,451--67\,469, 2022.

\bibitem{siciliano2009robotics}
B.~Siciliano, L.~Sciavicco, L.~Villani, and G.~Oriolo, \emph{Robotics}.\hskip
  1em plus 0.5em minus 0.4em\relax Springer London, 2009.

\bibitem{diankov2010automated}
R.~Diankov, ``Automated construction of robotic manipulation programs,'' Ph.D.
  dissertation, Carnegie Mellon University, The Robotics Institute Pittsburgh,
  2010.

\bibitem{diankov2008openrave}
R.~Diankov and J.~Kuffner, ``Openrave: A planning architecture for autonomous
  robotics,'' \emph{Robotics Institute, Pittsburgh, PA, Tech. Rep.
  CMU-RI-TR-08-34}, vol.~79, 2008.

\bibitem{hawkins2013analytic}
K.~P. Hawkins, ``Analytic inverse kinematics for the universal robots
  ur-5/ur-10 arms,'' \emph{Georgia Institute of Technology, Tech. Rep}, 2013.

\bibitem{hauser2020continuous}
K.~Hauser, ``Continuous pseudoinversion of a multivariate function: Application
  to global redundancy resolution,'' in \emph{Algorithmic Foundations of
  Robotics XII: Proceedings of the Twelfth Workshop on the Algorithmic
  Foundations of Robotics}.\hskip 1em plus 0.5em minus 0.4em\relax Springer,
  2020, pp. 496--511.

\bibitem{hemami1987more}
A.~Hemami, ``A more general closed-form solution to the inverse kinematics of
  mechanical arms,'' \emph{Advanced robotics}, vol.~2, no.~4, pp. 315--325,
  1987.

\bibitem{hollerbach1985optimum}
J.~M. Hollerbach, ``Optimum kinematic design for a seven degree of freedom
  manipulator,'' in \emph{Robotics research: The second international
  symposium}.\hskip 1em plus 0.5em minus 0.4em\relax Citeseer, 1985, pp.
  215--222.

\bibitem{shimizu2008analytical}
M.~Shimizu, H.~Kakuya, W.-K. Yoon, K.~Kitagaki, and K.~Kosuge, ``Analytical
  inverse kinematic computation for 7-dof redundant manipulators with joint
  limits and its application to redundancy resolution,'' \emph{IEEE
  Transactions on robotics}, vol.~24, no.~5, pp. 1131--1142, 2008.

\bibitem{faria2018position}
C.~Faria, F.~Ferreira, W.~Erlhagen, S.~Monteiro, and E.~Bicho, ``Position-based
  kinematics for 7-dof serial manipulators with global configuration control,
  joint limit and singularity avoidance,'' \emph{Mechanism and Machine Theory},
  vol. 121, pp. 317--334, 2018.

\bibitem{he2021analytical}
Y.~He and S.~Liu, ``Analytical inverse kinematics for franka emika panda--a
  geometrical solver for 7-dof manipulators with unconventional design,'' in
  \emph{2021 9th International Conference on Control, Mechatronics and
  Automation (ICCMA)}.\hskip 1em plus 0.5em minus 0.4em\relax IEEE, 2021, pp.
  194--199.

\bibitem{singh2010analytical}
G.~K. Singh and J.~Claassens, ``An analytical solution for the inverse
  kinematics of a redundant 7dof manipulator with link offsets,'' in \emph{2010
  IEEE/RSJ International Conference on Intelligent Robots and Systems}.\hskip
  1em plus 0.5em minus 0.4em\relax IEEE, 2010, pp. 2976--2982.

\bibitem{lavalle1998rapidly}
S.~M. LaValle \emph{et~al.}, ``Rapidly-exploring random trees: A new tool for
  path planning,'' 1998.

\bibitem{kavraki1996probabilistic}
L.~E. Kavraki, P.~Svestka, J.-C. Latombe, and M.~H. Overmars, ``Probabilistic
  roadmaps for path planning in high-dimensional configuration spaces,''
  \emph{IEEE transactions on Robotics and Automation}, vol.~12, no.~4, pp.
  566--580, 1996.

\bibitem{kuffner2000rrt}
J.~J. Kuffner and S.~M. LaValle, ``Rrt-connect: An efficient approach to
  single-query path planning,'' in \emph{Proceedings 2000 ICRA. Millennium
  Conference. IEEE International Conference on Robotics and Automation.
  Symposia Proceedings (Cat. No. 00CH37065)}, vol.~2.\hskip 1em plus 0.5em
  minus 0.4em\relax IEEE, 2000, pp. 995--1001.

\bibitem{bohlin2000path}
R.~Bohlin and L.~E. Kavraki, ``Path planning using lazy prm,'' in
  \emph{Proceedings 2000 ICRA. Millennium conference. IEEE international
  conference on robotics and automation. Symposia proceedings (Cat. No.
  00CH37065)}, vol.~1.\hskip 1em plus 0.5em minus 0.4em\relax IEEE, 2000, pp.
  521--528.

\bibitem{jaillet2010sampling}
L.~Jaillet, J.~Cort{\'e}s, and T.~Sim{\'e}on, ``Sampling-based path planning on
  configuration-space costmaps,'' \emph{IEEE Transactions on Robotics},
  vol.~26, no.~4, pp. 635--646, 2010.

\bibitem{karaman2011sampling}
S.~Karaman and E.~Frazzoli, ``Sampling-based algorithms for optimal motion
  planning,'' \emph{The international journal of robotics research}, vol.~30,
  no.~7, pp. 846--894, 2011.

\bibitem{ko2014randomized}
I.~Ko, B.~Kim, and F.~C. Park, ``Randomized path planning on vector fields,''
  \emph{The International Journal of Robotics Research}, vol.~33, no.~13, pp.
  1664--1682, 2014.

\bibitem{salzman2016asymptotically}
O.~Salzman and D.~Halperin, ``Asymptotically near-optimal rrt for fast,
  high-quality motion planning,'' \emph{IEEE Transactions on Robotics},
  vol.~32, no.~3, pp. 473--483, 2016.

\bibitem{otte2016rrtx}
M.~Otte and E.~Frazzoli, ``Rrtx: Asymptotically optimal single-query
  sampling-based motion planning with quick replanning,'' \emph{The
  International Journal of Robotics Research}, vol.~35, no.~7, pp. 797--822,
  2016.

\bibitem{zucker2013chomp}
M.~Zucker, N.~Ratliff, A.~D. Dragan, M.~Pivtoraiko, M.~Klingensmith, C.~M.
  Dellin, J.~A. Bagnell, and S.~S. Srinivasa,
  ``\href{https://journals.sagepub.com/doi/abs/10.1177/0278364913488805}{Chomp:
  Covariant hamiltonian optimization for motion planning},'' \emph{The
  International Journal of Robotics Research}, vol.~32, no. 9-10, pp.
  1164--1193, 2013.

\bibitem{kalakrishnan2011stomp}
M.~Kalakrishnan, S.~Chitta, E.~Theodorou, P.~Pastor, and S.~Schaal,
  ``\href{https://ieeexplore.ieee.org/abstract/document/5980280}{STOMP:
  Stochastic trajectory optimization for motion planning},'' in \emph{2011 IEEE
  international conference on robotics and automation}.\hskip 1em plus 0.5em
  minus 0.4em\relax IEEE, 2011, pp. 4569--4574.

\bibitem{toussaint2017tutorial}
M.~Toussaint, ``A tutorial on newton methods for constrained trajectory
  optimization and relations to slam, gaussian process smoothing, optimal
  control, and probabilistic inference,'' \emph{Geometric and numerical
  foundations of movements}, pp. 361--392, 2017.

\bibitem{petersen2023growing}
M.~Petersen and R.~Tedrake, ``Growing convex collision-free regions in
  configuration space using nonlinear programming,'' \emph{arXiv preprint
  arXiv:2303.14737}, 2023.

\bibitem{marcucci2023fast}
T.~Marcucci, P.~Nobel, R.~Tedrake, and S.~Boyd, ``Fast path planning through
  large collections of safe boxes,'' \emph{arXiv preprint arXiv:2305.01072},
  2023.

\bibitem{fernandez2018generative}
E.~Fern{\'a}ndez~Gonz{\'a}lez \emph{et~al.}, ``Generative multi-robot task and
  motion planning over long horizons,'' Ph.D. dissertation, Massachusetts
  Institute of Technology, 2018.

\bibitem{deits2015efficient}
R.~Deits and R.~Tedrake, ``Efficient mixed-integer planning for uavs in
  cluttered environments,'' in \emph{2015 IEEE international conference on
  robotics and automation (ICRA)}.\hskip 1em plus 0.5em minus 0.4em\relax IEEE,
  2015, pp. 42--49.

\bibitem{gottlieb1988topology}
D.~H. Gottlieb, ``Topology and the robot arm,'' \emph{Acta Applicandae
  Mathematica}, vol.~11, pp. 117--121, 1988.

\bibitem{burdick1989inverse}
J.~W. Burdick, ``On the inverse kinematics of redundant manipulators:
  Characterization of the self-motion manifolds,'' in \emph{Advanced Robotics:
  1989: Proceedings of the 4th International Conference on Advanced Robotics
  Columbus, Ohio, June 13--15, 1989}.\hskip 1em plus 0.5em minus 0.4em\relax
  Springer, 1989, pp. 25--34.

\bibitem{luck1993self}
C.~L. Luck and S.~Lee, ``Self-motion topology for redundant manipulators with
  joint limits,'' in \emph{[1993] Proceedings IEEE International Conference on
  Robotics and Automation}.\hskip 1em plus 0.5em minus 0.4em\relax IEEE, 1993,
  pp. 626--631.

\bibitem{cohn2023noneuclidean}
T.~Cohn, M.~Petersen, M.~Simchowitz, and R.~Tedrake, ``{Non-Euclidean Motion
  Planning with Graphs of Geodesically-Convex Sets},'' in \emph{Proceedings of
  Robotics: Science and Systems}, Daegu, Republic of Korea, July 2023.

\bibitem{lee2012smooth}
J.~M. Lee and J.~M. Lee, \emph{Smooth manifolds}.\hskip 1em plus 0.5em minus
  0.4em\relax Springer, 2012.

\bibitem{snopt}
P.~E. Gill, W.~Murray, and M.~A. Saunders, ``{SNOPT}: An {SQP} algorithm for
  large-scale constrained optimization,'' \emph{SIAM Rev.}, vol.~47, pp.
  99--131, 2005.

\bibitem{tedrake2023underactuated}
\BIBentryALTinterwordspacing
R.~Tedrake, \emph{Underactuated Robotics}, 2023. [Online]. Available:
  \url{https://underactuated.csail.mit.edu}
\BIBentrySTDinterwordspacing

\bibitem{tedrake2019drake}
\BIBentryALTinterwordspacing
R.~Tedrake and the Drake Development~Team, ``Drake: Model-based design and
  verification for robotics,'' 2019. [Online]. Available:
  \url{https://drake.mit.edu}
\BIBentrySTDinterwordspacing

\bibitem{schwarzer2004exact}
F.~Schwarzer, M.~Saha, and J.-C. Latombe, ``Exact collision checking of robot
  paths,'' \emph{Algorithmic foundations of robotics V}, pp. 25--41, 2004.

\bibitem{mosek}
M.~ApS, ``\href{https://www.mosek.com/documentation/}{MOSEK Optimization
  Suite},'' 2019.

\bibitem{sucan2012ompl}
I.~A. {\c{S}}ucan, M.~Moll, and L.~E. Kavraki, ``The {O}pen {M}otion {P}lanning
  {L}ibrary,'' \emph{{IEEE} Robotics \& Automation Magazine}, vol.~19, no.~4,
  pp. 72--82, December 2012, \url{https://ompl.kavrakilab.org}.

\end{thebibliography}

\end{document}